\pdfoutput=1

\documentclass[11pt]{article}

\usepackage[dvipsnames]{xcolor} 

\usepackage{authblk}
\usepackage{acl}
\usepackage{times}
\usepackage{latexsym}
\usepackage[T1]{fontenc}
\usepackage{graphicx}
\usepackage{float} 
\usepackage{amssymb}
\usepackage{amsthm}
\usepackage{booktabs}
\usepackage{adjustbox}

\usepackage[utf8]{inputenc}

\usepackage{microtype}

\title{BenCoref: A Multi-Domain Dataset of Nominal Phrases and Pronominal Reference Annotations}

\author{
Shadman Rohan$^{1}$, Mojammel Hossain$^{1}$, Mohammad Mamun Or Rashid$^{2}$, Nabeel Mohammed$^{1}$ \\
$^{1}$North South University \quad $^{2}$Jahangirnagar University \\
{\tt{shadman.rohan, mojammel.hossain, nabeel.mohammed}@northsouth.edu} \\
{\tt{mamunbd@juniv.edu}} \
}

\begin{document}
\maketitle
\begin{abstract}

Coreference Resolution is a well studied problem in NLP. While widely studied for English and other resource-rich languages, research on coreference resolution in Bengali largely remains unexplored due to the absence of relevant datasets. Bengali, being a low-resource language, exhibits greater morphological richness compared to English. In this article, we introduce a new dataset, BenCoref, comprising coreference annotations for Bengali texts gathered from four distinct domains. This relatively small dataset contains 5200 mention annotations forming 502 mention clusters within 48,569 tokens. We describe the process of creating this dataset and report performance of multiple models trained using BenCoref. We expect that our work provides some valuable insights on the variations in coreference phenomena across several domains in Bengali and encourages the development of additional resources for Bengali. Furthermore, we found poor crosslingual performance at zero-shot  setting from English, highlighting the need for more language-specific resources for this task. The dataset is available at \footnote{codes used to generate the results along with data is available at: \url{https://github.com/ShadmanRohan/BenCoref}}.
\end{abstract}

\section{Introduction}


Coreference resolution is the task of identifying all references to the same entity in a document. This task originally started as a sub-task of information extraction. The Message Understanding Conferences \cite{grishman1996message} first introduced three tasks, collectively referred to as SemEval, designed to measure the deeper understanding of any information extraction (IE) system. One of these three tasks proposed in the event was coreferencial noun phrase identification. 

The Automatic Content Extraction (ACE) Program \cite{doddington-etal-2004-automatic} was the first major initiative that created a large dataset with entity, event and relation annotations. This project revealed some major complexities behind creating such dataset. Some of the significant challenges reported by the annotators include the coreference of generic entities, use of metonymy, characterization of Geo-Political Entity, distinguishing certain complex relations, and recognizing implicit vs. explicit relations.

\begin{figure}
    \centering
    \includegraphics[width=0.9\columnwidth]{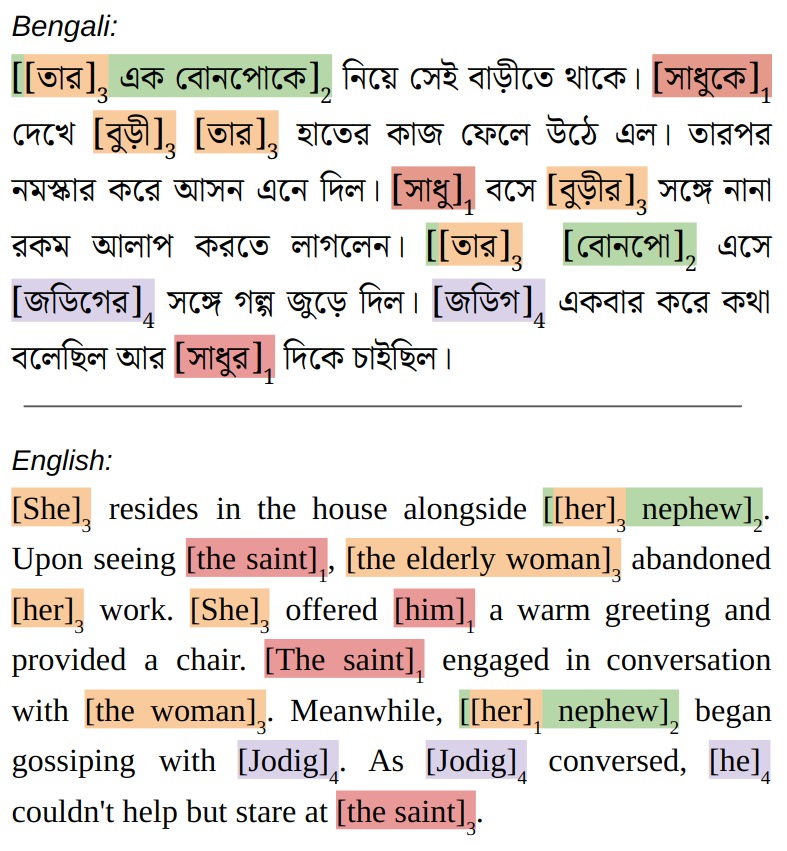}
    \caption{BenCoref annotations with  color-coded Co-reference chains.
}
    \label{fig:sample}
\end{figure}

Since then coreference resolution, anaphoric \& cataphoric relation identification, event reference detection has been studied widely. As a result, large datasets like ACE \cite{doddington-etal-2004-automatic}, Ontonotes \cite{pradhan2012conll}, WikiCoref \cite{ghaddar2016wikicoref}, and LitBank \cite{bamman-etal-2020-annotated} were made public. Some datasets, like ACE \cite{doddington-etal-2004-automatic}, and Ontonotes, expanded this task beyond English to include more languages, like Arabic, and Chinese. 

This coreference resolution task has shown potential in improving many downstream NLP tasks like machine translation \cite{miculicich2017using, ohtani2019context}, literary analysis \cite{bamman-etal-2014-bayesian}, question anwering \cite{morton1999using}, text summarization \cite{steinberger2007two}, etc. However, Bengali, despite being a popular lanaguage, has seen very little work is this direction due to lack of public datasets.  

Figure \ref{fig:sample} shows a sample from our dataset with each color representing an unique entity. The main contributions of this work are: 

\begin{itemize}
\item We introduce a new Bengali coreference annotated dataset, consisting of 48,569 tokens collected from four diverse domains. Our dataset creation process is shared along with the annotators' guidelines, which we believe is the first of its kind for Bengali coreference annotation.

\item We characterize the behaviour and distribution of nominal and pronominal coreference mentions across the four domains with necessary statistics. Furthermore, we report the performance of an end-to-end neural coreference resolution system that was solely built using our data.

\item We empirically demonstrate the necessity for more language-specific datasets, particularly for low-resource languages, by comparing our results with zero-shot cross-lingual learning from English.

\end{itemize}

\subsection{Related Datasets}

To the best of our knowledge, no coreference dataset in Bengali exists. Most of the works related to Bengali \cite{sikdar2013adapting, senapati2013guitar, sikdar2015differential} uses data from ICON2011 shared task which was never publicly shared.

Most of the major coreference datasets are in English. OntoNotes \cite{pradhan2012conll} is a well-annotated and large dataset with over 1.6M words. This dataset does not contain any singleton mention. Later, LitBank \cite{bamman-etal-2020-annotated} was published that is almost 10 times larger than OntoNotes (12.3M words).

\section{Challenges in Bengali}
One of the main challenges we faced was the absence of preexisting coreference annotation guidelines tailored for the Bengali language. To overcome this obstacle, we adapted the OntoNotes coreference annotation guideline to suit our objectives. This highlighted several distinctive linguistic characteristics of Bengali, such as zero anaphora, non-anaphoric pronouns, and case-marking, that needs to be carefully considered when preforming co-reference annotation in Bengali. Each of this is discussed with more details and examples in Figure \ref{bengaliCharacteristics} in the Appendix.

Moreover, we discovered that existing annotation software is ill-equipped to manage Bengali text, occasionally leading to inaccurate rendering and unstable character display. This underscores the importance of advancing normalization techniques and standardization of Bengali digital representation.

\section{Data Domain Description}
The Bengali language can be braoadly categorized into two primary literary dialects, namely "Shadhubhasha" and "Choltibhasha." "Shadhubhasha" was commonly used by Bangla writers and individuals in the 19th and early 20th centuries, while "Choltibhasha" is currently the more prevalent and colloquial dialect. This dataset contains both domains of Bengali text, with story and novel texts sourced from copyright-free books of the 19th and 20th centuries, and biography and descriptive texts obtained from modern sources, primarily in "Choltibhasha."  A brief description of each domain is given below:

\subsection{Biography}
A biography presents a comprehensive account of an individual's life, character, accomplishments, and works, spanning from birth to death or the present time. Although the number of references per document in biographical texts is comparable to other genres, they primarily focus on a single subject throughout the entire narrative. Additionally, the dialect employed in biographies in BenCoref is typically "Choltibhasha."

\subsection{Descriptive}
By descriptive text we refer to wikipedia-like articles. They cover a broad range of subjects that span various fields, such as technology, professions, travel, economics, and numerous related subtopics. These comprehensive texts try to accurately portray and convey holistic information about real-world objects or experiences.

\subsection{Story}
BenCoref is primarily composed of short stories, each with a word count of 1000 words or less, which was an arbitrary decision. These stories typically feature 3-4 characters on average. The language used in the stories varies, with some being exclusively in "Shadhubhasha," while others use a mix of "Shadhubhasha" and "Choltibhasha."

\subsection{Novel}
The Bengali novels in our dataset typically consist of more than 1200 words and feature an average of over 5 characters. These novels primarily employ "Shadhubhasha".
The next segment discusses the coreference behaviour across each domain in more detail.

\section{Domain Specific Coreference Behaviour Characterization}
In this section, some statistics is presented to better understand the coreference phenomenon across each domain. Each coreference cluster may refer to different type of entities, like an object, people, location or event. An arbitrary design choice was made to not explicitly mark the type of entity.

We start by analyzing the mean and standard deviation between mentions across the domains. Table \ref{tab:std} shows that biographies and novels exhibit a low standard deviation but have noticeably different mean distance between mentions. On the other hand, stories and descriptive texts fall in the middle, exhibiting a similar coreference distribution. For mentions that span more than one token, only the first token was used for calculation.

\begin{table}[htp]
\centering
\begin{tabular}{ccc}
\hline
\textbf{Categories}	& \textbf{Mean}	& \textbf{Std. Dev}\\
\hline
Novel &29.17 & 3.70 \\
Story &24.10 & 8.46 \\
Biography &15.67 & 3.81 \\
Descriptive &22.35 & 5.42 \\
\hline
\end{tabular}
\caption{\label{tab:std} Mean and Std. Deviation of distance between mentions in each domain.}
\label{tab:accents}
\end{table}

\begin{figure}[H]
\includegraphics[width=\columnwidth]{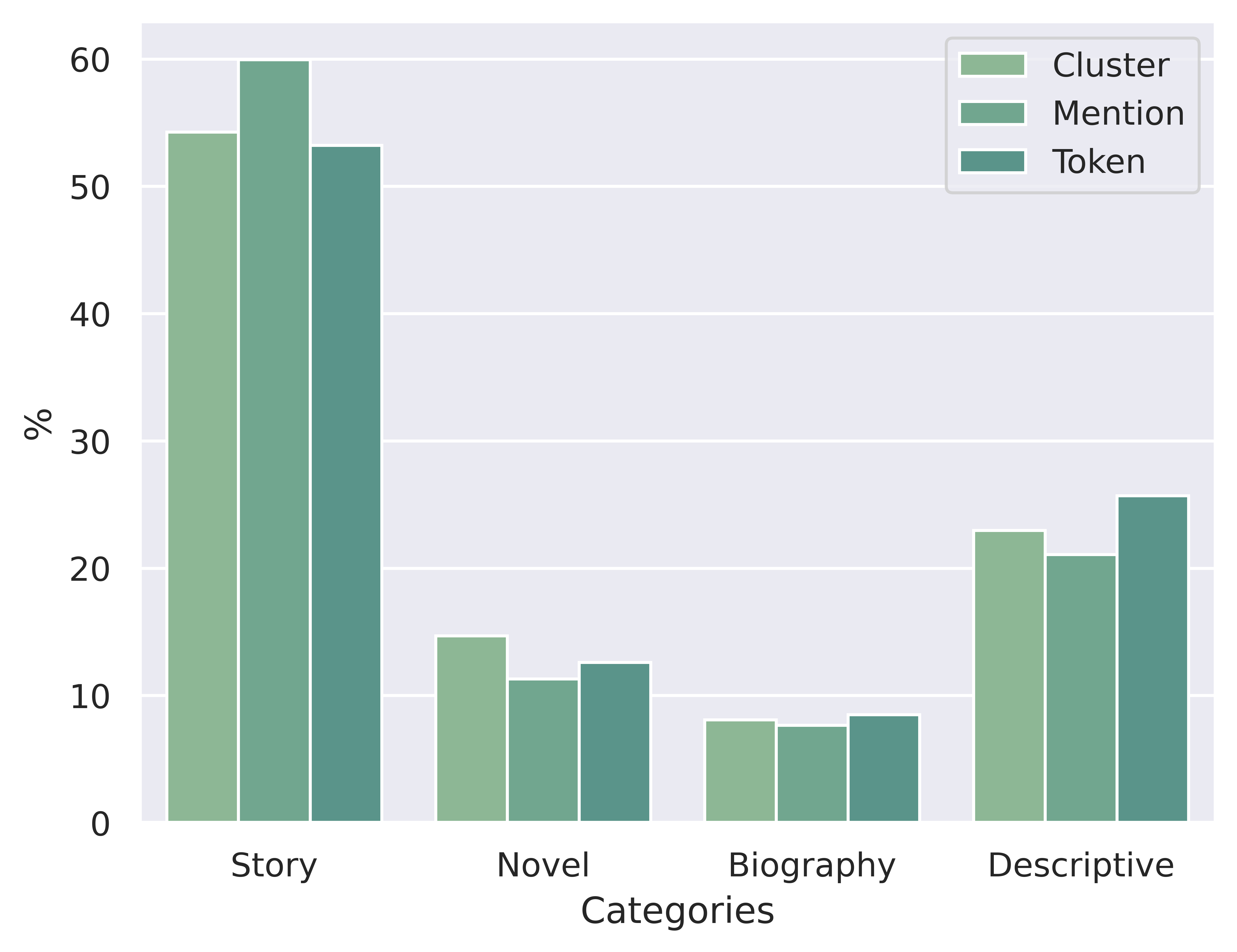}
\caption{(Right) Distribution of Clusters, Mentions, and Tokens across the categories.}\label{fig:dist}
\end{figure}

The majority of texts in BenCoref belong to the stories domain, while the biography domain has the smallest contribution. The distribution of mentions, clusters, and tokens across the categories in BenCoref is presented in Figure \ref{fig:dist}.

\begin{figure}[H]
\includegraphics[width=1\columnwidth]{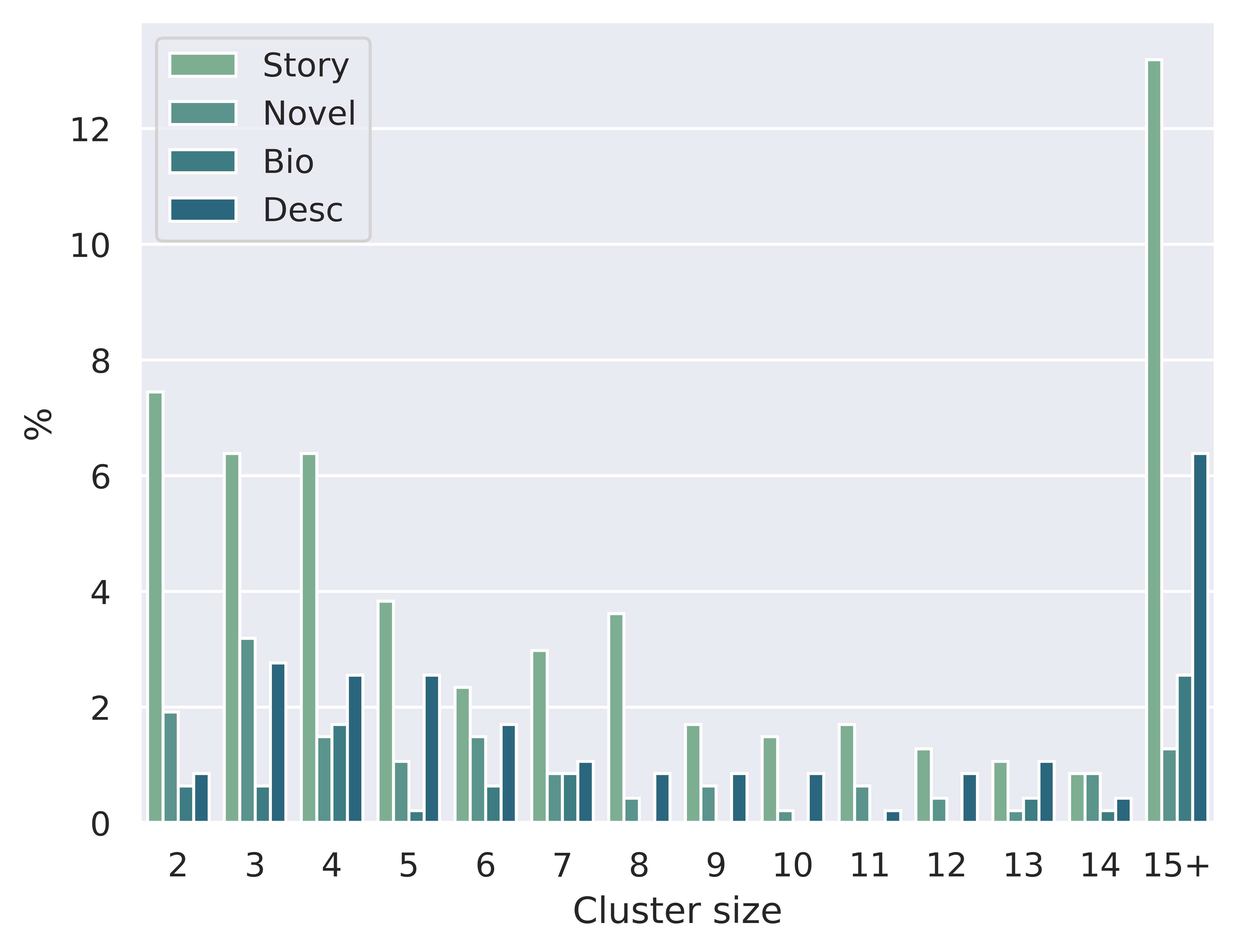}
\caption{Cluster size comparison between Story, Novel, Biography and wiki-like Descriptive domain.}\label{fig:cluster}
\end{figure}

Figure \ref{fig:cluster} depicts the distribution of cluster size across each domain. The cluster size refers to the total number of mentions in each coreference chain. It is worth noting that singletons were not annotated in BenCoref. The story domain has the highest number of coreference chains with two mentions only. Since the story domain contributes the most data to the dataset, this may be a contributing factor to its high frequency in each cluster size. Besides story, the descriptive domain also seems to have more larger coreference chains.

\begin{figure}[H]
\includegraphics[width=\columnwidth]{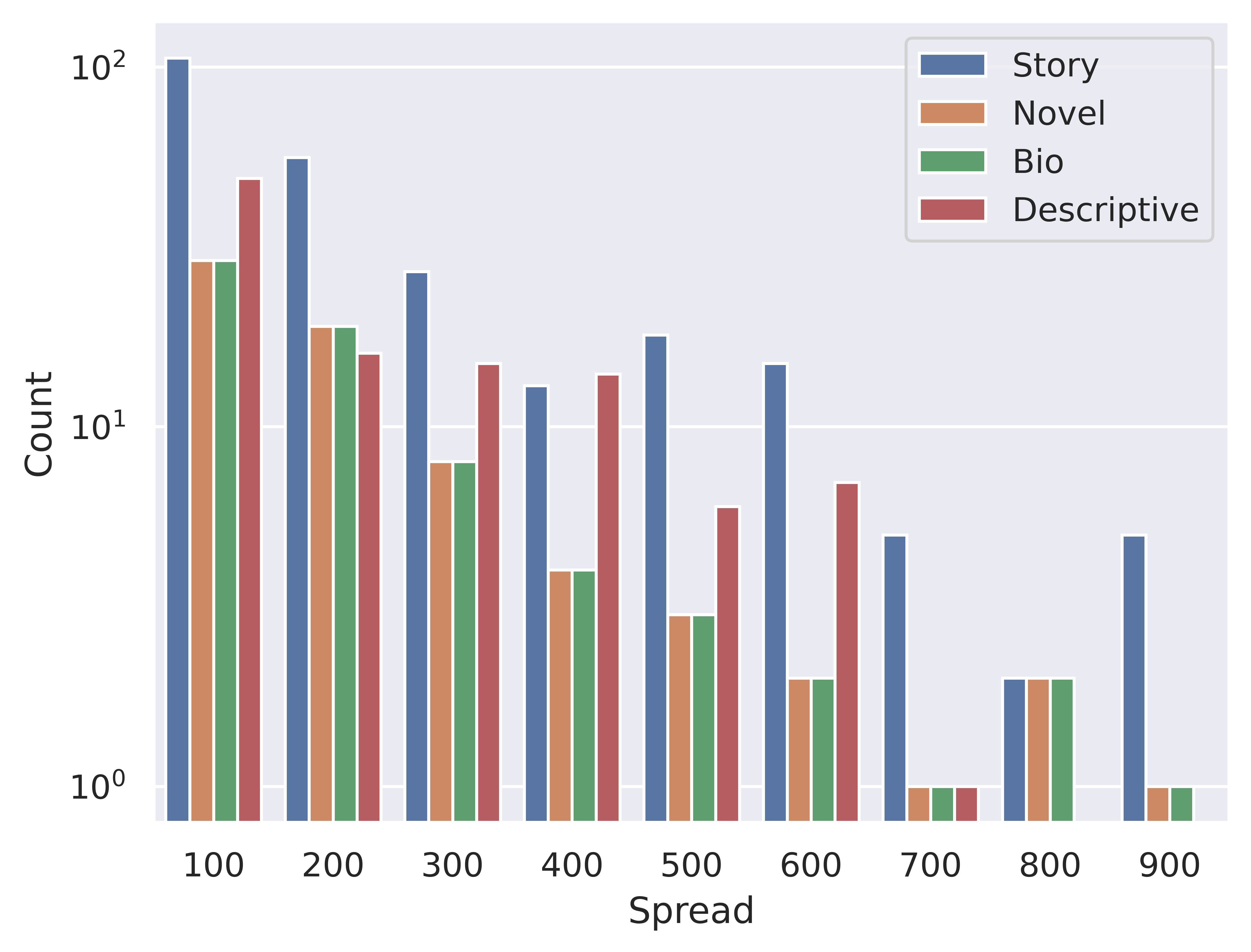}
\caption{Spread in BenCoref across each domain. The spread is measured by the token level distance between the first and last mention of an entity.}\label{fig:spread}
\end{figure}

Figure \ref{fig:spread} compares the spread of coreference chains in each domain, where the spread refers to the token-level distance between the beginning and end of a coreference chain. A general trend can be observed that as the size of the coreference chain increases, its corresponding frequency decreases in each domain.

An additional "index.csv" file is included with in the dataset, which serves as an index to all the documents included, organized by title and author. A partial view of this file is presented in Appendix Figure \ref{fig:index_csv}.

\section{Methodology}

We used BnWikiSource\footnote{\url{https://bn.wikipedia.org}} and Banglapedia\cite{islam2003banglapedia} as sources of copyright-free Bengali text for our dataset. Banglapedia was used for biographies and wiki-like descriptive texts. The dataset creation process is discussed in detail in the following paragraphs.

\subsection{Annotation Phase}
The WebAnno annotator \cite{eckart-de-castilho-etal-2016-web} was the chosen tool for annotation. To accommodate WebAnno's limited capacity to work with large texts, the articles were partitioned according to the Table \ref{tab:doc}. Each partition ends in a complete sentence and any incomplete portion of a sentence were moved to the next fragment. The partition size was chosen arbitrarily to reduce the number of data fragments. In Appendix Figure {\ref{fig:annotation_ss}}, a screenshot of the WebAnno interface used during this phase is displayed. A post annotation sample is provided in Figure \ref{fig:bio_gold} in Appendix from the Biography domain.

\begin{table}[htp]
\centering
\begin{tabular}{ccc}
\hline
\textbf{Tokens}	& \textbf{Partitions}\\
\hline
<699&1\\
<1000&2\\
>1000&3\\
\hline
\end{tabular}
\caption{\label{tab:doc} Documents with greater than 699 tokens and less than 1000 tokens were divided into 2 parts and the ones with more than 1000 tokens were divided into 3.}
\end{table}

Since there is no existing guideline for coreference annotation, the annotators were initially instructed to annotate the noun phrases and its coreferences, which were predominantly pronouns. The primary noun phrase references were tagged as "entity" and their corresponding coreferences were tagged as "ref". While determining what forms an entity is an important linguistic problem, it is not the primary challenge we are trying to address in our work. Annotators were free to mark any token or span that they considered an entity. After the annotation phase was completed, the data was exported and the character-level annotations were converted to token-level annotations. For every exceptional cases encountered, a new rule was established and enforced during further annotation of the dataset. The rules are further discussed in the next section.

\subsection{Annotation Strategy/Guideline}
This coreference annotation guide (refer to \ref{rules} in the Appendix) was prepared concurrently with the annotation phase to ensure consistency throughout the annotation process. We mirrored the overall structure of the OntoNotes annotation guidelines, tailoring them to our specific use case.

Initially, we did not impose any specific restrictions on the definition of an entity during the annotation process. The annotators were instructed to annotate any span they deemed as an entity. However, this approach resulted in an annotator bias, with a strong focus on nominal and pronominal mentions. Subsequently, we made the decision to prioritize and concentrate solely on these types of mentions.

Furthermore, as part of our design decision, we chose to not tag singleton mentions. Consequently, any singletons were removed during the post-annotation processing phase.

\subsection{Annotation Criteria:}

The general rule used for annotation is to annotate mentions in any form, including nested mentions or those referring to multiple entities. The characterization of mention and coreference link types was conducted after annotating the entire dataset. Annotating coreference link types was kept optional due to the significant training required for the task. This strategy was followed the accelerate the annotation process. 

The rules with corresponding examples are illustrated in a more detailed manner in Figure \ref{guide} in the Appendix. Furthermore, the coreference link types have been categorized into two groups, namely identical and apposite, and they have been discussed in detail in \ref{identical} and \ref{appositive} in the Appendix. However, the task of annotating coreference link types is currently pending and will be addressed in future work.

While this guideline is incomplete and limited in scope, it can play an impotant role  in encouraging the next generation of coreference datasets in Bengali. The OntoNotes coreference guideline\cite{guideline} is currently in its 7th edition which is a strong indication that the first attempt on making a such guideline would be imperfect and will require further revisions. It may take several iterations before we can have a robust guideline for coreference annotation in Bengali.

\subsection{Inter-Annotator Agreement}

The OntoNotes strategy was roughly employed to assess interannotator agreement in this work. Specifically, two annotators independently annotated the documents, and only in cases of disagreement, a third annotator was consulted to arrive at a final decision. These ultimate annotations were deemed as the gold standard annotations.

Based on the adjudicated version as the ground truth, the individual annotations in our dataset achieved an average MUC score of 78.3 on the combined dataset.  while the combined inter-annotator MUC score was 67.6.

However, it is important to acknowledge that the process of resolving disagreements was not adequately documented and will be addressed in greater detail in future endeavors.

\section{Experiments} 
We took an end-to-end neural network based modeling approach. The following section discusses the algorithm, followed by the experimental setup, evaluation strategy and analysis of results.

We used the 300-dimensional Fasttext and Glove embeddings \cite{Fasttext} as words representations. To generate contexual representations the embeddings were passed through a bi-directional LSTM \cite{hochreiter1997lstm} for some experiments and a variation of the popular transformer-based \cite{vaswani2017attention} pretrained model, BERT\cite{devlin-etal-2019-bert}, for other experiments. For the task of coreference resolution, the contextual representations from these base models were passed on to a span ranking model-head, originally proposed in \cite{lee-etal-2018-higher}. 
For the crosslingual experiment, a multilingual BERT was finetuned on the OntoNotes dataset.

For hyperparameter optimization, we tuned the maximum number of words in a span(s), maximum number of antecedents per span(a), and coref layer depth(CL). 

\subsection{Experimental Setup}
The data was separated into train and dev set on a ratio of 95\% by 5\%. An additional test set was carefully prepared, completely disjoint from the train and dev set, that contains 37 documents. An overview of the dataset given in Table \ref{tab:dataset}

\begin{table}[htp]
\footnotesize
\centering
\begin{tabular}{ll c c c}
    \toprule
    &\textbf{categories}& \textbf{documents} &\textbf{mentions} &\textbf{clusters}\\
    \toprule
    train  &biography &17 &421 &38\\
    + dev &descriptive &36 &1157 &108\\
    &novel &13 &601 &78\\
    &story &56 &3021 &278\\
    \midrule
    test &biography &10 &303 &22\\
    &descriptive &9 &290 &33\\
    &novel &3 &191 &15\\
    &story &15 &697 &53\\
    \bottomrule
\end{tabular}
\caption{\label{tab:dataset}Dataset distribution}
\end{table}

For evaluating our system, we used the CONLL-2012 official evaluation scripts which calculates four metrics: Identification of Mentions, MUC, B3 and CEAF. The following section analyzes the performance of our model.

\subsection{Results and Analysis}

\begin{table}[!htbp]
\centering
\begin{adjustbox}{width=1 \columnwidth}
\begin{tabular}{lllccc}
\toprule
\textbf{category}& \textbf{model} &\textbf{parameters} &pre. &rec. &f1\\
\midrule
&c2f+Glove &s=30, a=50, CL=2 &93.83 &65.34 &77.04 \\
biography&c2f+Fasttext &s=20, a=50, CL=2 &{\colorbox{pink}{96.51}} &64.02 &76.98\\
&BERT-base &s=30, a=50, CL=2 &94.22 &{\colorbox{pink}{86.13}} &{\colorbox{pink}{90.00}}\\
&M-BERT(Zero-Shot) &s=30, a=50, CL=2 &7.14 &4.62 &5.61\\
\midrule
&c2f+Glove &s=30, a=50, CL=2 &73.78 &58.96 &65.55 \\
story&c2f+Fasttext &s=20, a=50, CL=2 &74.80 &54.08 &62.78\\
&BERT-base &s=30, a=50, CL=2 &{\colorbox{pink}{83.91}} &{\colorbox{pink}{65.85}} &{\colorbox{pink}{73.79}}\\
&M-BERT(Zero-Shot) &s=30, a=50, CL=2 &7.40 &3.73 &4.96\\
\midrule
&c2f+Glove &s=30, a=50, CL=2 &78.00 &40.83 &53.60 \\
novel&c2f+Fasttext &s=20, a=50, CL=2 &{\colorbox{pink}{87.50}} &43.97 &58.53\\
&BERT-base &s=30, a=50, CL=2 &85.41 &{\colorbox{pink}{64.39}} &{\colorbox{pink}{73.43}}\\
&M-BERT(Zero-Shot) &s=30, a=50, CL=2 &8.51 &4.18 &5.61\\
\midrule
&c2f+Glove &s=30, a=50, CL=2 &66.39 &27.93 &39.32 \\
descriptive&c2f+Fasttext &s=20, a=50, CL=2 &72.16 &24.13 &36.17\\
&BERT-base &s=30, a=50, CL=2 &{\colorbox{pink}{82.95}} &{\colorbox{pink}{50.34}} &{\colorbox{pink}{62.66}}\\
&M-BERT(Zero-Shot) &s=30, a=50, CL=2 &7.47 &5.51 &6.34\\
\bottomrule
\end{tabular}
\renewcommand\thetable{3}
\end{adjustbox}
\caption{\label{tab:identscore}Identification of mentions}
\end{table}

\begin{table*}[!htbp]
\centering
\begin{adjustbox}{width={\textwidth},totalheight={\textheight},keepaspectratio}%
\begin{tabular}{lllcccccccccccc}
\toprule
& & & & $B^3$ & & & MUC & & & $CEAF_{\phi4}$ & & &Avg &\\
\toprule
category& &parameters &Pre. &Rec. &F1 &Pre. &Rec. &F1 &Pre. &Rec. &f1 &Pre. &Rec. & F1\\
\midrule
&c2f + Glove &s=30, a=50, CL=2 &84.52 &44.74 &58.51 &92.26 &64.41 &76.05 &55.33 &40.24 &46.60 &77.37 &49.80 &60.39\\

biography&c2f + Fasttext &s=20, a=50, CL=2 &{\colorbox{pink}{89.09}} &43.99 &58.90 &{\colorbox{pink}{95.74}} &64.05 &76.75 &{\colorbox{pink}{61.24}} &36.19 &45.49 &{\colorbox{pink}{82.02}} &48.08 &60.38\\

&BERT-base &s=30, a=50, CL=2 &85.37 &{\colorbox{pink}{73.59}} &{\colorbox{pink}{79.04}} &93.79 &{\colorbox{pink}{86.12}} &{\colorbox{pink}{89.79}} &57.48 &{\colorbox{pink}{49.64}} &{\colorbox{pink}{53.27}} &78.88 &{\colorbox{pink}{69.78}} &{\colorbox{pink}{74.03}}\\
&M-BERT(Zero-Shot) &s=30, a=50, CL=2 &4.28 &0.25 &0.46 &0.67 &0.35 &0.46 &1.22 &2.93 &1.72 &2.06 &1.18 &0.88\\

\midrule
&c2f + Glove &s=30, a=50, CL=2 &46.92 &23.95 &31.72 &63.41 &44.40 &52.23 &20.24 &36.99 &26.16 &43.52 &35.11 &36.70\\

story&c2f + Fasttext &s=20, a=50, CL=2 &47.31 &22.80 &30.77 &65.23 &42.54 &51.50 &23.49 &34.02 &27.79 &45.34 &33.12 &36.69\\

&BERT-base &s=30, a=50, CL=2 &{\colorbox{pink}{54.64}} &{\colorbox{pink}{31.62}} &{\colorbox{pink}{40.06}} &{\colorbox{pink}{74.46}} &{\colorbox{pink}{53.88}} &{\colorbox{pink}{62.52}} &{\colorbox{pink}{28.80}} &{\colorbox{pink}{40.23}} &{\colorbox{pink}{33.57}} &{\colorbox{pink}{52.63}} &{\colorbox{pink}{41.91}} &{\colorbox{pink}{45.38}}\\

&M-BERT(Zero-Shot) &s=30, a=50, CL=2 &2.32 &0.25 &0.45 &1.42 &0.62 &0.86 &2.00 &2.59 &2.26 &1.91 &1.15 &1.19\\

\midrule
&c2f + Glove &s=30, a=50, CL=2 &49.87 &7.98 &13.77 &59.45 &25.00 &35.20 &16.37 &26.61 &20.27 &41.90 &19.86 &23.08\\

novel&c2f + Fasttext &s=20, a=50, CL=2 &{\colorbox{pink}{60.30}} &10.45 &17.82 &{\colorbox{pink}{72.72}} &31.81 &44.26 &23.36 &27.74 &25.37 &{\colorbox{pink}{52.13}} &23.33 &29.15\\

&BERT-base &s=30, a=50, CL=2 &43.55 &{\colorbox{pink}{33.93}} &{\colorbox{pink}{38.14}} &71.31 &{\colorbox{pink}{52.27}} &{\colorbox{pink}{60.32}} &{\colorbox{pink}{34.98}} &{\colorbox{pink}{32.80}} &{\colorbox{pink}{33.85}} &49.95 &{\colorbox{pink}{39.67}} &{\colorbox{pink}{44.10}}\\

&M-BERT(Zero-Shot) &s=30, a=50, CL=2 &3.54 &0.25 &0.47 &2.66 &1.13 &1.59 &1.90 &2.49 &2.15 &2.70 &1.29 &1.40\\

\midrule
&c2f + Glove &s=30, a=50, CL=2 &48.33 &11.91 &19.12 &58.24 &20.62 &30.45 &31.16 &26.83 &28.83 &45.91 &19.79 &26.13\\

descriptive&c2f + Fasttext &s=20, a=50, CL=2 &58.32 &9.74 &16.70 &66.66 &18.67 &29.17 &29.98 &20.81 &24.57 &51.65 &16.41 &23.48\\

&BERT-base &s=30, a=50, CL=2 &{\colorbox{pink}{62.81}} &{\colorbox{pink}{26.88}} &{\colorbox{pink}{37.65}} &{\colorbox{pink}{76.62}} &{\colorbox{pink}{45.91}} &{\colorbox{pink}{57.42}} &{\colorbox{pink}{46.12}} &{\colorbox{pink}{28.18}} &{\colorbox{pink}{34.99}} &{\colorbox{pink}{61.85}} &{\colorbox{pink}{33.66}} &{\colorbox{pink}{43.35}}\\

&M-BERT(Zero-Shot) &s=30, a=50, CL=2 &2.01 &0.56 &0.88 &1.16 &0.77 &0.93 &2.30 &3.00 &2.60 &1.82 &1.44 &1.47\\
\bottomrule
\end{tabular}
\end{adjustbox}
\caption{Performance on test data. The main evaluation metric is the average F1 score of $MUC$, $B^3$, and $CEAF_{\phi 4}$. The best scores are highlighted.}
\label{tab:clusterscore}
\end{table*}

The performance of the model was reasonable given the size of our dataset. As neural networks tend to achieve optimal performance with larger datasets, we hypothesize that our results could be enhanced by expanding our dataset. The model demonstrated good performance in identifying individual mentions, as evidenced by the scores presented in Table \ref{tab:identscore}. However, we observed a decrease in performance during the second phase of clustering the mentions, as shown in Table \ref{tab:clusterscore}. This highlights the challenge of accurately identifying coreference clusters, particularly in languages with complex sentence structures and a high degree of lexical ambiguity. Further innovation is needed to address these challenges and improve the overall performance of coreference resolution models.

Upon closer inspection one recurring problem was discovered. The model failed to do basic common sense reasoning on long coreference clusters, often breaking it up into several clusters. As demonstrated in Figure \ref{fig:bio_pred} in Appendix, the model failed to merge clusters 0 and 1, which should have been a single cluster.

Furthermore, it can be observed that the coreference resolution model performs significantly better on the biography domain as compared to other domains. The relatively low mean and standard deviation of the distance between mentions reported in Table \ref{tab:std} may have contributed to this result. However, despite forming the major portion of the dataset, the story domain did not show any significant improvement. The high standard deviation in distance between mentions reported in Table \ref{tab:std} for the story domain may have contributed to this lack of improvement. Qualitative analysis is needed to investigate the underlying causes of this performance gap.

The zero-shot crosslingual experiment demostrated that coreference knowledge doesn't easily transfer through multilingual training. This clearly demonstrates the need for language specific datasets. Some studies \cite{novak2014cross} report developing projection techniques to improve crosslingual coreference resolution. There maybe scope for further work in this direction.
\section{Conclusion}

This paper presented BenCoref, the first publicly available dataset of coreference annotations in Bengali. The creation process and annotation guidelines were described in detail to facilitate future work in this area. We then used the dataset to develop an end-to-end coreference resolution system and reported its performance across different domains. Our findings indicate that a lower mean and standard deviation of token-distance between mentions may lead to better results, but further experiments on other datasets are needed to confirm this hypothesis. We also observed a higher tendency for breakage in longer coreference chains.

Our zero-shot cross-lingual experiment demonstrated that coreference knowledge does not easily transfer through multilingual training, highlighting the importance of language-specific datasets. While some studies \cite{novak2014cross} have reported success in developing projection techniques to improve cross-lingual coreference resolution, further research is required to explore this area.

\section*{Acknowledgements}
We would like to express our sincere gratitude to everyone who contributed to this research. We would also like to thank our colleagues and collaborators for their valuable insights and feedback throughout the project. We are especially grateful to our annotators for their hard work and dedication to ensure the quality of this dataset. We also appreciate the feedback and suggestions from the anonymous reviewers, which helped to improve the quality of this manuscript. No financial assistance was provided by any organization for the completion of this project.

\bibliography{anthology,custom}
\bibliographystyle{acl_natbib}

\appendix

\section{Appendices}

\label{rules}

\begin{table*}[!htbp]
\centering
\begin{adjustbox}{width={\textwidth},totalheight={\textheight},keepaspectratio}%
\begin{tabular}{lllcccccccccccc}
\toprule
& & & & $B^3$ & & & MUC & & & $CEAF_{\phi4}$ & & &Avg &\\
\toprule
Category& &Parameters &Pre. &Rec. &F1 &Pre. &Rec. &F1 &Pre. &Rec. &F1 &Pre. &Rec. & F1\\
\midrule
&c2f + Fasttext &s=30, a=250, CL=2 &93.43 &45.94 &61.59 &97.88 &65.83 &78.72 &65.62 &41.76 &51.04 &85.64 &51.18 &63.78\\

Biography&c2f + Fasttext &s=20, a=50, CL=3 &91.63 &45.20 &60.54 &96.84 &65.48 &78.13 &65.87 &41.92 &51.23 &84.78 &50.87 &63.30\\

&c2f + Fasttext &s=10, a=50, CL=3 &90.98 &47.81 &62.68 &96.92 &67.25 &79.41 &77.44 &42.24 &54.66 &88.54 &52.44 &65.58\\

\midrule
&c2f + Fasttext &s=30, a=250, CL=2 &48.07 &22.95 &31.07 &65.78 &38.81 &48.82 &21.98 &33.73 &26.62 &45.28 &31.83 &35.50\\

Story&c2f + Fasttext &s=20, a=50, CL=3 &53.08 &21.70 &30.81 &67.02 &38.50 &48.91 &22.17 &34.02 &26.84 &47.42 &31.41 &35.52\\

&c2f + Fasttext &s=10, a=50, CL=3 &53.82 &19.35 &28.47 &66.76 &34.62 &45.60 &21.92 &34.77 &26.89 &47.50 &29.58 &33.65\\

\midrule
&c2f + Fasttext &s=30, a=250, CL=2 &47.06 &7.32 &12.67 &61.03 &26.70 &37.15 &16.04 &23.06 &18.92 &41.38 &19.03 &22.91\\

Novel&c2f + Fasttext &s=20, a=50, CL=3 &65.10 &8.54 &15.10 &71.42 &25.56 &37.65 &24.96 &28.08 &26.42 &53.83 &20.73 &26.39\\

&c2f + Fasttext &s=10, a=50, CL=3 &57.71 &5.82 &10.57 &63.79 &21.02 &31.62 &17.03 &23.42 &19.72 &46.18 &16.75 &20.64\\

\midrule
& c2f + Fasttext &s=30, a=250, CL=2 &56.43 &9.88 &16.82 &61.84 &18.28 &28.22 &25.72 &20.01 &22.51 &48.00 &16.06 &22.52\\

Descriptive&c2f + Fasttext &s=20, a=50, CL=3 &53.78 &7.64 &13.37 &58.92 &12.84 &21.08 &28.76 &19.17 &23.01 &47.15 &13.22 &19.15\\

&c2f + Fasttext &s=10, a=50, CL=3 &52.41 &7.46 &13.06 &58.06 &14.00 &22.57 &24.47 &19.03 &21.41 &44.98 &13.50 &19.01\\

\bottomrule
\end{tabular}
\end{adjustbox}
\renewcommand\thetable{2}
\caption{Some additional results on the Model's performance.}
\label{tab:allscore}
\end{table*}

\begin{table}[!htbp]
\centering
\begin{adjustbox}{width=1 \columnwidth}
\begin{tabular}{lllccc}
\toprule
Category& &Parameters &Pre. &Rec. &F1\\
\midrule
&c2f+Fasttext &s=30, a=250, CL=2 &98.52 &66.00 &79.05\\
Biography&c2f+Fasttext &s=20, a=50, CL=3 &97.54 &65.67 &78.50\\
&c2f+Fasttext &s=10, a=50, CL=3 &98.52 &66.00 &79.05\\
\midrule
&c2f+Fasttext &s=30, a=250, CL=2 &74.20 &49.92 &59.69\\
Story&c2f+Fasttext &s=20, a=50, CL=3 &75.16 &49.49 &59.68\\
&c2f+Fasttext &s=10, a=50, CL=3 &76.52 &46.77 &58.05\\
\midrule
&c2f+Fasttext &s=30, a=250, CL=2 &79.00 &41.36 &54.29\\
Novel&c2f+Fasttext &s=20, a=50, CL=3 &90.12 &38.21 &53.67\\
&c2f+Fasttext &s=10, a=50, CL=3 &83.75 &35.07 &49.44\\

\midrule
&c2f+Fasttext &s=30, a=250, CL=2 &74.03 &26.55 &39.08\\
Descriptive&c2f+Fasttext &s=20, a=50, CL=3 &70.00 &19.31 &30.27\\
&c2f+Fasttext &s=10, a=50, CL=3 &68.88 &21.37 &32.63\\
\bottomrule
\end{tabular}
\renewcommand\thetable{3}
\end{adjustbox}
\caption{\label{tab:identscore_additional} Additional identification of mention results}
\end{table}

\begin{figure*}
  \includegraphics[width=\textwidth]{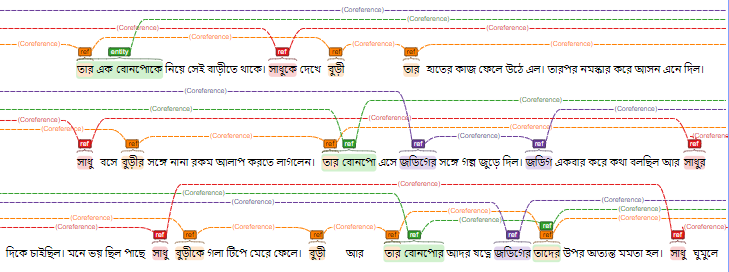}
  \caption{A screenshot from WebAnno\cite{eckart-de-castilho-etal-2016-web} during annotation phase. In this example, the highlighted words are marked as mentions and same color indicate the mentions belong to the same cluster. A colored line joins the highlighted words creating a chain forming a single cluster.}\label{fig:annotation_ss}
\end{figure*}

\begin{figure*}
    \centering
    \includegraphics[width=13cm]{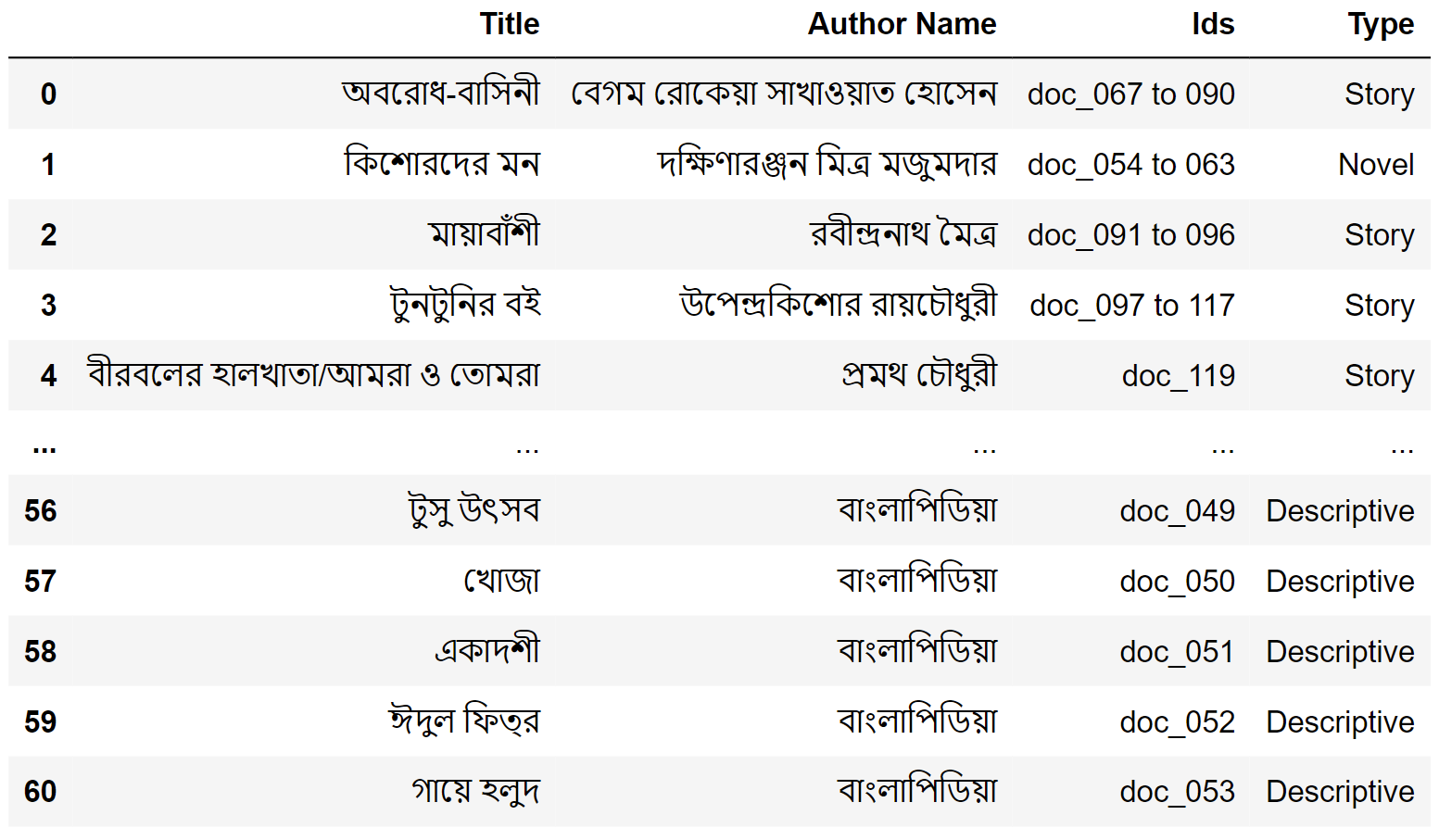}
    \caption{The supplementary datafile index.csv contains an index to all the datapoints}\label{fig:index_csv}
    \label{index}
\end{figure*}

\begin{figure*}[!htbp]
    \centering
    \includegraphics[width=1\textwidth]{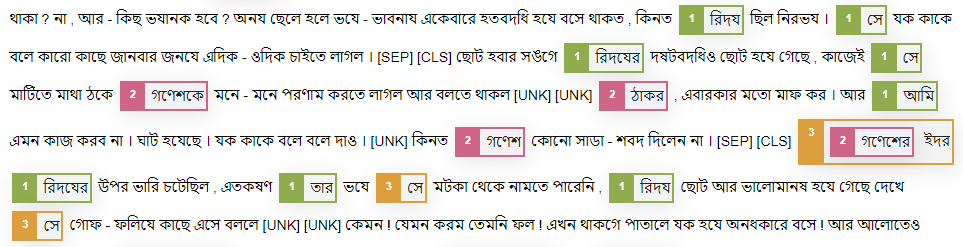}
    \caption{Biography document}
    \label{fig:bio_gold}
\end{figure*}

\begin{figure*}[!htb]
    \centering
    \includegraphics[width=1\textwidth]{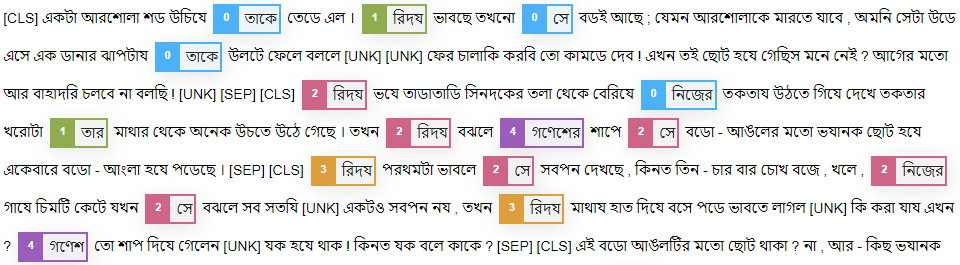}
    \caption{BERT-base model's prediction on a biography document}
    \label{fig:bio_pred}
\end{figure*}

\begin{figure*}
    \centering
    \includegraphics[width=16cm]{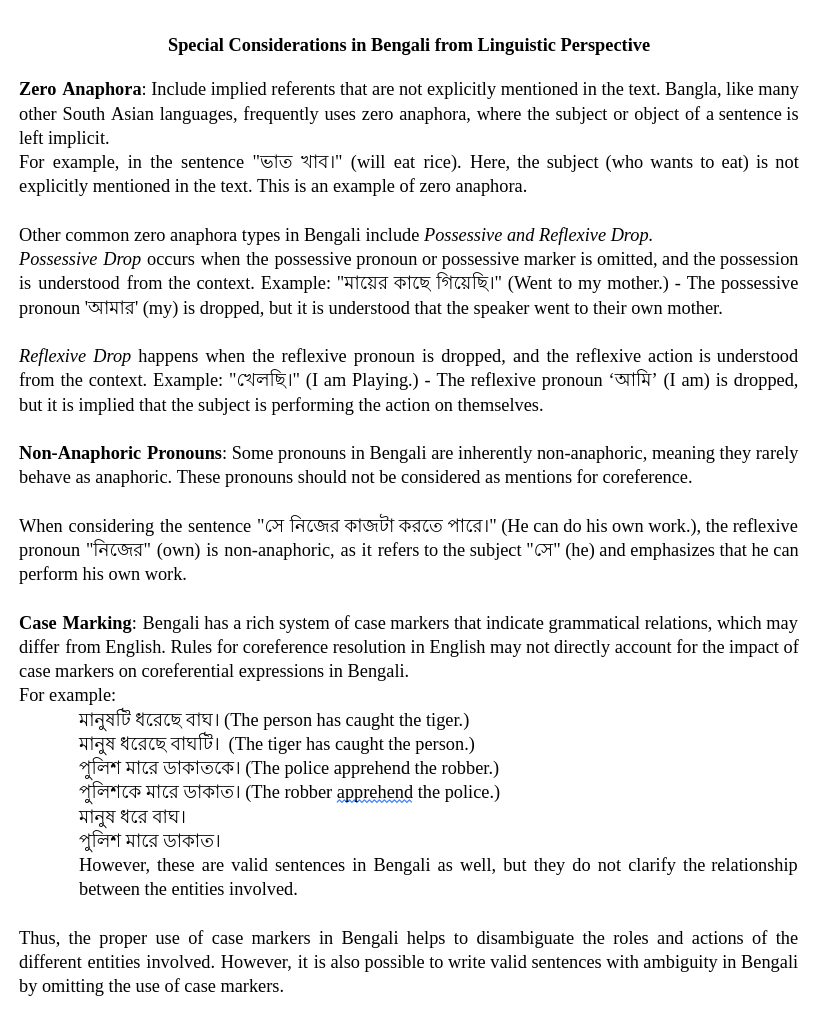}
    \caption{This highlights few key considerations when annotating coreference in Bengali.}
    \label{bengaliCharacteristics}
\end{figure*}

\begin{figure*}
    \centering
    \includegraphics[width=16cm]{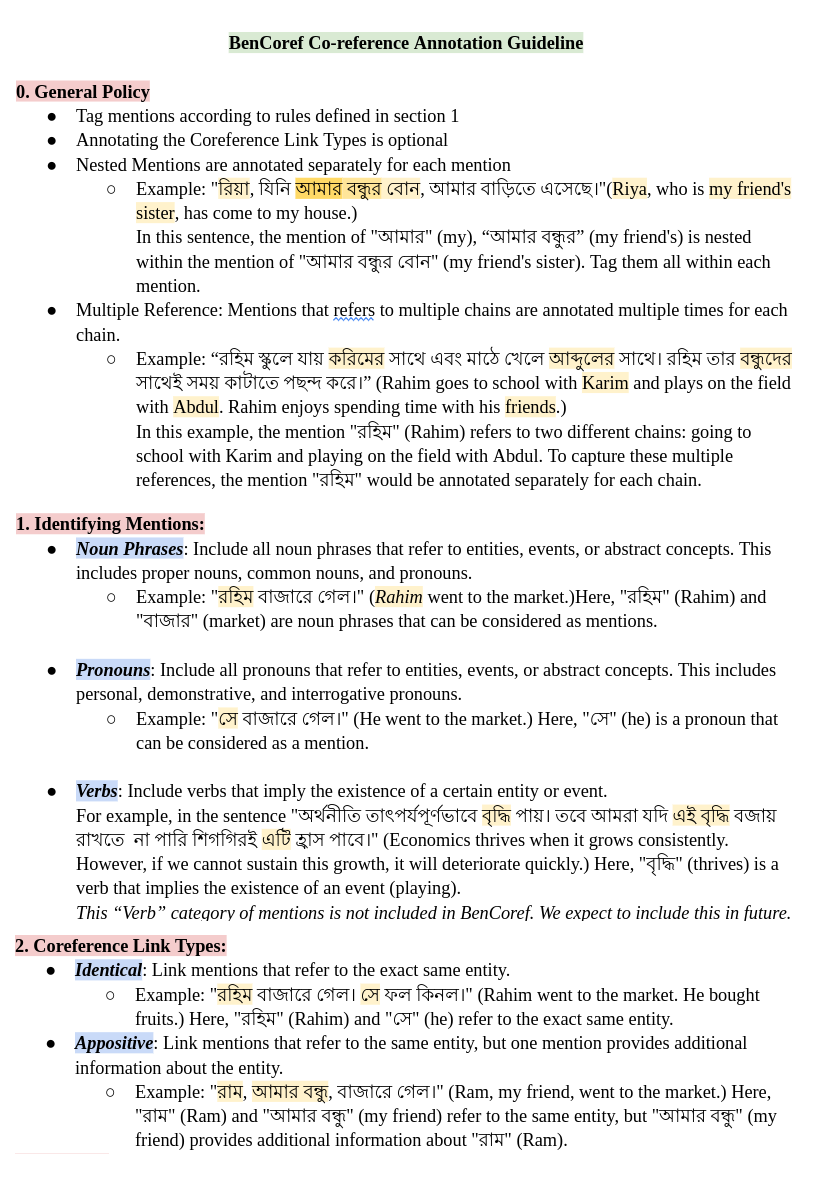}
    \caption{BenCoref Annotation Guideline with examples.}
    \label{guide}
\end{figure*}

\begin{figure*}
    \centering
    \includegraphics[width=16cm]{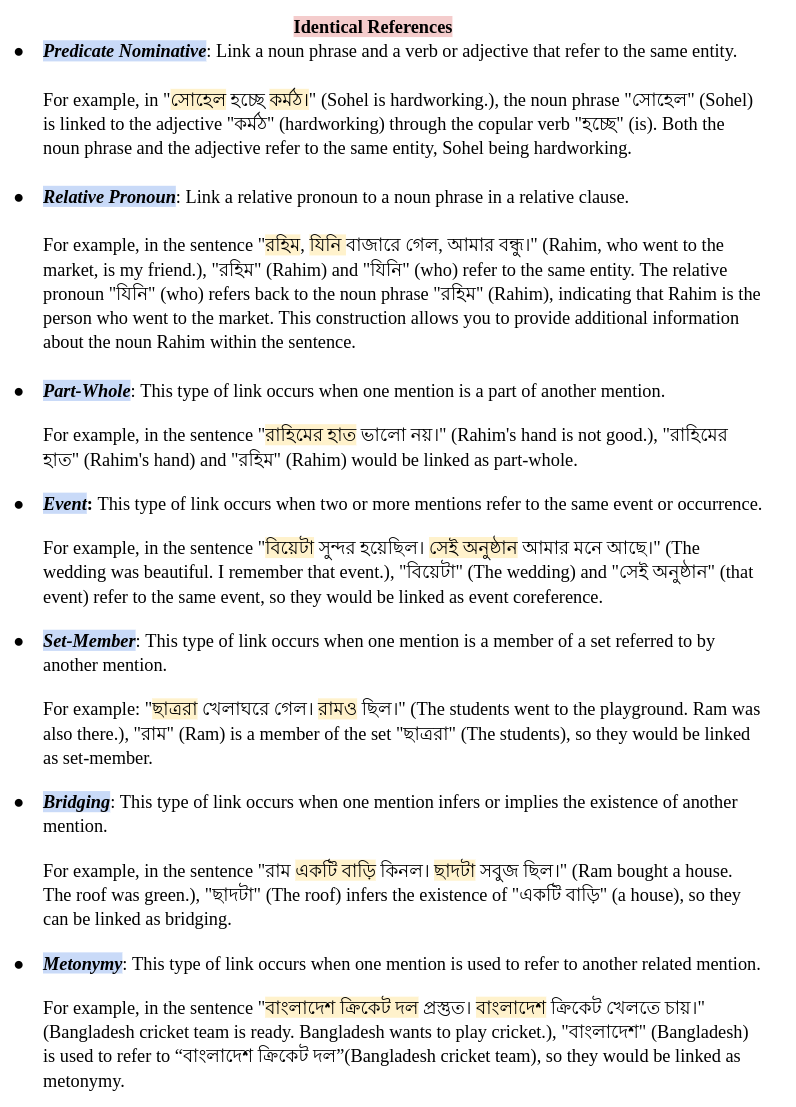}
    \caption{Identical Reference Types}
    \label{identical}

\end{figure*}

\begin{figure*}
    \centering
    \includegraphics[width=16cm]{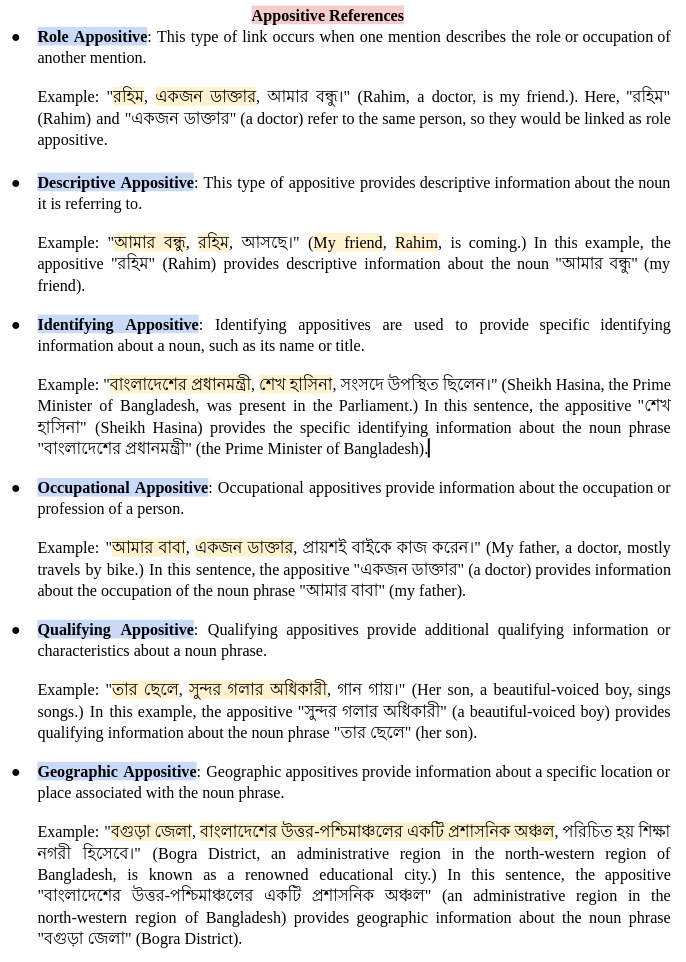}
    \caption{Appositive Reference Types}
    \label{appositive}
\end{figure*}

\end{document}